\documentclass[german,english]{2554}

\title{Comparative Analysis of Unsupervised Algorithms for Breast MRI Lesion Segmentation}

\titlerunning{Unsupervised Algorithms for Breast MRI Lesion Segmentation}

\author{Sulaiman Vesal$^1$, Nishant Ravikumar$^1$, Stephan Ellman$^2$, Andreas Maier$^1$}
\authorrunning{Vesal et al.}

\institute{%
	$^1$Fakult\"at f\"ur Pattern Recognition, FAU Erlangen-N\"urnberg, Germany
	$^2$Radiologisches Institut, Universit\"atsklinikum Erlangen, Germany
}

\email{sulaiman.vesal@fau.de}

\begin{document}

%
\selectlanguage{english}

\maketitle

\begin{abstract}
Accurate segmentation of breast lesions is a crucial step in evaluating the characteristics of tumors. However, this is a challenging task, since breast lesions have sophisticated shape, topological structure, and variation in the intensity distribution. In this paper, we evaluated the performance of three unsupervised algorithms for the task of breast Magnetic Resonance (MRI) lesion segmentation, namely, Gaussian Mixture Model clustering, K-means clustering and a marker-controlled Watershed transformation based method. All methods were applied on breast MRI slices following selection of regions of interest (ROIs) by an expert radiologist and evaluated on 106 subjects' images, which include 59 malignant and 47 benign lesions. Segmentation accuracy was evaluated by comparing our results with ground truth masks, using the Dice similarity coefficient (DSC), Jaccard index (JI), Hausdorff distance and precision-recall metrics. The results indicate that the marker-controlled Watershed transformation outperformed all other algorithms investigated. 
\end{abstract}

\section{Introduction}
Breast cancer is one of the leading causes of mortality in women \cite{XI2017145}, which can be significantly reduced through early detection and treatment. Breast lesions found during screening examinations are more likely to be smaller and still confined to the breast. There are different imaging modalities for breast cancer screening like mammograms which can find breast changes before symptom development. Ultrasound is also useful to differentiate between cysts and solid masses in women with dense breast tissues. However, in current clinical practice,  Magnetic Resonance (MR) images of the breast, are assessed visually or using basic quantitative measures such as lesion diameter and apparent diffusion coefficient (from diffusion-weighted MRIs). Breast cancer diagnosis and distinguishing malignant from benign tumors is infeasible using such measures due to low precision \cite{XI2017145}. Accurate lesion segmentation is a crucial step in evaluating tumor characteristics and addressing these limitations. This is a challenging task as, lesions boundaries are usually obscured, irregular, have low contrast and overlap with healthy tissue \cite{JMRI:JMRI24394}\cite{THOMASSINNAGGARA2012828}. 

In this study, we investigate three unsupervised methods for lesion segmentation in breast MRIs, namely, Gaussian mixture model (GMM) clustering, K-means (KM) clustering and marker-controlled Watershed transformation (MCWT), and compare their performance. The primary advantage of unsupervised methods over supervised ones is that they do not require ground truth segmentations, which are cumbersome and time-consuming to evaluate manually for radiologists. Additionally, such manual segmentations are intrinsically subjective and hence tend to vary between raters. This attribute also makes unsupervised methods uniquely suitable for automatic segmentation in real-time \cite{ZHANG2008260}\cite{univis91680126}.

\section{Methods and Materials}

\subsection{K-Means Clustering}
K-means (KM) clustering is a simple unsupervised algorithm for image segmentation. The procedure follows an easy way to partition $n$ pixels into $k$ clusters. Each pixel is assigned to a cluster with the closest mean. The mean of each cluster is often referred to as the centroid and pixels assigned to a cluster are more similar than those assigned to other clusters. The algorithm iteratively alternates between assigning pixels to a cluster, based on their distance to cluster centroids, and refining estimates for the cluster centroids \cite{Moftah:2014:AKC:2628703.2628727}. This is achieved by minimizing the mean-squared-error objective function $S(V)$ given by: 

\begin{equation}
\label{mse}
S(V) = \sum_{i=1}^{c_i}\sum_{j=1}^{c_j}(||x_i-v_j||)^2
\end{equation}

where, $||x_i-v_j||$ is the Euclidean distance between pixel $x_i$ and mean $v_j$, $c_i$ is the number of pixels and $c_j$ is the number of cluster centers. Equation \ref{mse} indicates that KM clustering is sensitive to the initial cluster assignment and the choice of the distance measure. We initialized the centroids through iterating over all pixels, find the distances between them and those with the largest distance considered as the initial centroids.

\subsection{Gaussian Mixture Model Clustering}
A Gaussian Mixture Model (GMM) is a parametric probability density function represented as a weighted sum of Gaussian component densities \cite{MP7483}. GMM parameters are estimated from training data using the Expectation-Maximization (EM) through an iterative process. Let us consider a 2D MR image $I$ as a vector of $N$ pixel values $x={x_1,x_2,...,x_N}$ and assume that they are realizations of a $k$-component GMM. Given a class $k$, with parameters $\theta_k=\{\mu_k,\sigma_k\}$, the conditional probability of the $i$\textsuperscript{th} pixel is expressed as shown in equation \ref{gauss}. Assuming all pixels in an image are independent and identically distributed samples of a $K$ component GMM, their joint probability may be expressed as shown in equation \ref{jprob}. Here $\lbrace x_i \rbrace_{i=1...N} = X$ is a $2D$-dimensional continuous-valued image vector, $\lbrace \theta_k \rbrace_{k=1...K} = \Theta$ represents the set of all model parameters, $\lbrace w_k\rbrace_{k=1...K}$ are the mixture weights, and $\mathcal{N}(x_i|\mu_k,\sigma_k)$ are the component Gaussian densities, with mean $\mu_k$ and covariance $\sigma_k$.

\begin{subequations}
	\begin{equation}
	\label{gauss}
	p(x_i|\mu_k,\sigma_k) = \dfrac{1}{(2\pi)^\frac{D}{2}|\sigma_k|^\frac{1}{2}}\exp\{-\frac{1}{2}(x_i-\mu_k')\Sigma_k^{-1} (x_i-\mu_k)\},
	\end{equation}
	\begin{equation}
	\label{jprob}
	p(X|\Theta) =\prod^{N}_{i=1}\sum_{k=1}^{K} w_k \mathcal{N}(x_i|\mu_k,\sigma_k)
	\end{equation}
\end{subequations}

GMMs are thus parameterized by their mean vectors, covariance matrices and mixture weights of the constituent component densities. The parameters $\Theta$ are estimated using the expectation-maximization(EM) algorithm \cite{MP7483}, which iteratively maximizes the expected complete data likelihood by alternating between the (E)xpectation and (M)aximization steps. The M-step updates for each model parameter are evaluated as follows:

\begin{subequations}
	\begin{equation}
	\mu_k^{(t+1)}=\frac{\sum\limits_{i=1}^{N} P^{t}_{ik} x_i}{\sum\limits_{i=1}^{N} P^{t}_{ik}},
	\end{equation}
	\begin{equation}
	\sigma_k^{(t+1)}=\frac{\sum\limits_{i=1}^{N}\sum\limits_{k=1}^{K} P^{t}_{ik} ||x_{i} - \mu_k^{(t+1)}||^2}{D\sum\limits_{i=1}^{N}\sum\limits_{k=1}^{K}P^{t}_{ik}},
	\end{equation}
	\begin{equation}
	\pi_k^{(t+1)} = \frac{1}{N}\sum\limits_{i}^{N}P^{t}_{ik}
	\end{equation}
\end{subequations}

In these equations $P^{t}_{ik}$ represents the posterior probability estimated at the $t$\textsuperscript{th} EM-iteration (as shown in equation \ref{E-step}), using the current estimates for the model parameters and $D$ is the dimension of the data being clustered. 

\begin{equation}
\label{E-step}
P^{t}_{ik} = \frac{\pi_{k}\mathcal{N}(x_i|\mu_k,\sigma_k)}{\sum\limits_{k=1}^{K} \pi_{k}\mathcal{N}(x_i|\mu_k,\sigma_k)}
\end{equation}

The estimated posterior probabilities, in turn, represent the cluster membership of the image pixels and are used to assign pixels to distinct clusters/classes, thereby segmenting the image.

\subsection{Marker-Controlled Watershed Transformation}
In previous work \cite{Vesal2017}, we proposed a robust and novel Marker-Controlled Watershed Transformation (MCWT) for the task of breast MRI lesion segmentation. As a pre-processing step for the MCWT, we computed the morphological gradient of the image, which is the pointwise difference between a unitary dilation and erosion. The gradient image provides information about edges. Normally, there are several local minima in a gradient image due to inherent noise in the original image, and direct application of the watershed transformation generally results in over-segmentation \cite{7760610}. To prevent over-segmentation, we defined markers to guide the watershed algorithm. Each marker is considered to be part of a specific watershed region and after segmentation, the boundaries of the regions are arranged to separates each object.

In MR images, tumor region is brighter and has more uniform intensity than its surroundings, which makes a good candidate for watershed segmentation. Based on this fact, we determined the internal and external markers by sorting out the pixel values in ROIs in descending order and chose $n$ pixels with maximum intensity values as markers. To find the optimal number of markers for this dataset, we tested the algorithm by varying the number of markers between $1$ and $150$. We found 45 markers to be optimal based on the segmentation accuracy achieved. 

\subsection{Data Acquisition}
MR images for this study were acquired on 1.5 T scanners Magnetom Avanto and 3.0 T  Magnetom Verio, Siemens Healthineers, Erlangen, Germany, with dedicated breast array coils and the patient in a prone position. The contrast media was applied into the cubital vein after the first of six dynamic acquisitions with a flow of 1.0 mL/sec chased by a 20 mL saline flush. One hundred and six lesions were identified from a representative set of 80 female patients by two expert radiologists who have more than 7 years of experience in evaluation of clinical findings. The mean patient age was 50 $\pm$ 13  and in all cases, cancer status was confirmed using histopathology. 42 of the lesions were diagnosed as benign and the remaining 64 as malignant.

\subsection{Pre-Processing}
In this study, a 2D slice was picked from the T1-weighted subtraction MR volume manually, based on the ground truth segmentation (also in 2D) provided by radiologists. Subsequently, a regions of interest(ROI) was drawn around the lesions, ensuring that all lesions identified by the radiologist were completely covered, as in some cases there were several lesions present, scattered across the breast. Additionally, to enhance image contrast, we applied contrast-limited adaptive histogram equalization(CLAHE) \cite{Reza2004}. The evaluation of segmentation methods described above was conducted using Dice similarity coefficient (DSC), Jaccard index (JI), Hausdorff distance (HD), precision (PR) and recall (RE) metrics \cite{MP7483}\cite{Xu2011}.

\section{Experimental Results}
Table 1 summarizes the segmentation accuracy achieved using each method (evaluated in terms of five metrics), for 106 lesions. Dice Coefficient, Jaccard index, Hausdorff distance, precision, and recall values were evaluated with respect to the ground truth segmentations and averaged over all cases, for each algorithm. Table 1 indicates that MCWT achieved higher segmentation accuracy compared to the rest and fewer false positives and false negative. Fig.1 shows examples of segmentation for each algorithm, along with their corresponding ground truth. 

\begin{table}[t]
	\caption{DSC, JI and HD results $(mean\pm std)$ for the different algorithms.}
	\label{0000-tab-schriften}
	\begin{tabular*}{\textwidth}{l@{\extracolsep\fill}lllll}
		\hline
		Methods     & DSC            & JI              & HD(mm)                   & PR              & RE        \\ \hline
		K-Means     & 0.732$\pm$0.206 & 0.612$\pm$0.209 & 2.292$\pm$1.05 &    0.805$\pm$0.243    & 0.702$\pm$0.204    \\
		GMM         & 0.746$\pm$0.180 & 0.623$\pm$0.193 & 2.275$\pm$1.08 &    0.855$\pm$0.213    & 0.697$\pm$0.195\\
		MCWT         & 0.786$\pm$0.172 & 0.679$\pm$0.217 & 2.265$\pm$1.24 &0.866$\pm$0.199  & 0.752$\pm$0.250    \\ \hline
	\end{tabular*}
\end{table}

\begin{figure}[htb]
	\centering
	\subfigure{\includegraphics[height=5\baselineskip, width=6\baselineskip]{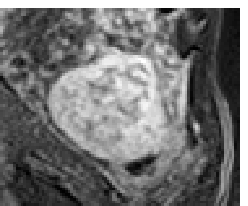}}\hfill%
	\subfigure{\includegraphics[height=5\baselineskip, width=6\baselineskip]{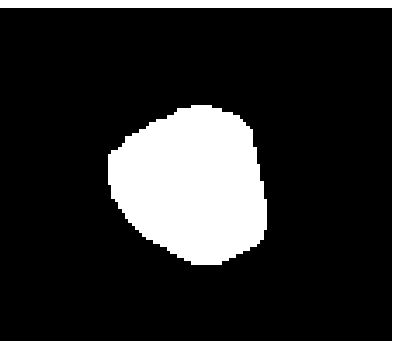}}\hfill%
	\subfigure{\includegraphics[height=5\baselineskip, width=6\baselineskip]{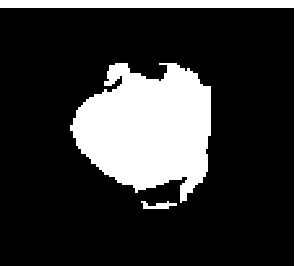}}\hfill%
	\subfigure{\includegraphics[height=5\baselineskip, width=6\baselineskip]{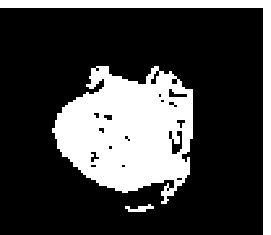}}\hfill%
	\subfigure{\includegraphics[height=5\baselineskip, width=6\baselineskip]{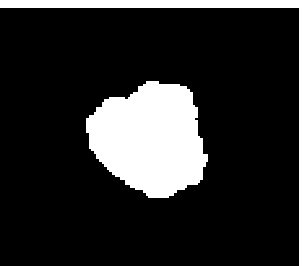}}\hfill%
	
	\subfigure{\includegraphics[height=5\baselineskip, width=6\baselineskip]{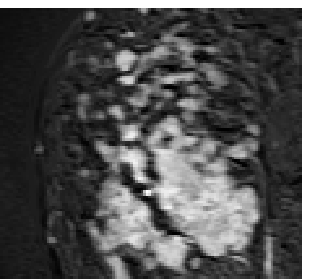}}\hfill%
	\subfigure{\includegraphics[height=5\baselineskip, width=6\baselineskip]{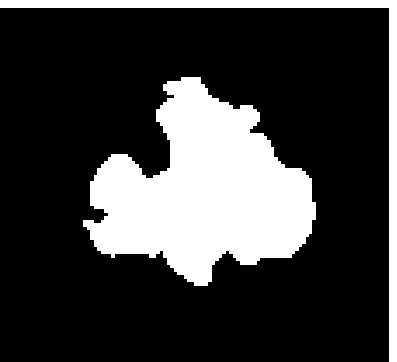}}\hfill%
	\subfigure{\includegraphics[height=5\baselineskip, width=6\baselineskip]{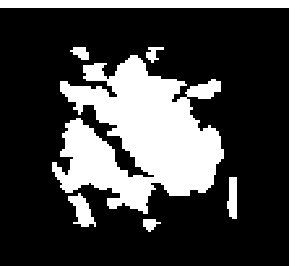}}\hfill%
	\subfigure{\includegraphics[height=5\baselineskip, width=6\baselineskip]{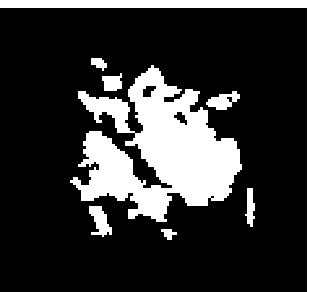}}\hfill%
	\subfigure{\includegraphics[height=5\baselineskip, width=6\baselineskip]{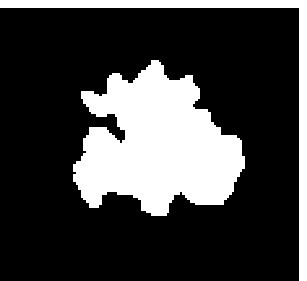}}\hfill%

	\setcounter{subfigure}{0}%
	\subfigure[Lesion]{\includegraphics[height=5\baselineskip, width=6\baselineskip]{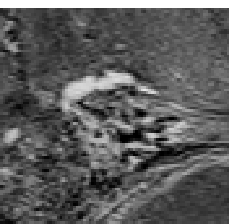}}\hfill%
	\subfigure[Mask]{\includegraphics[height=5\baselineskip, width=6\baselineskip]{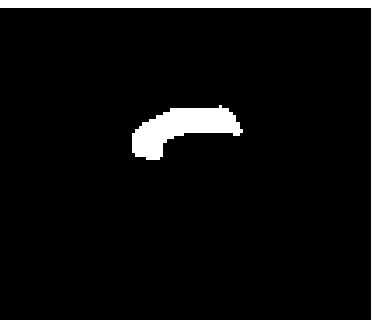}}\hfill%
	\subfigure[GMM]{\includegraphics[height=5\baselineskip, width=6\baselineskip]{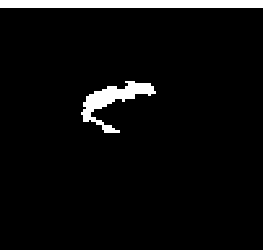}}\hfill%
	\subfigure[KM]{\includegraphics[height=5\baselineskip, width=6\baselineskip]{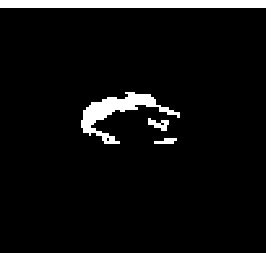}}\hfill%
	\subfigure[MCWT]{\includegraphics[height=5\baselineskip, width=6\baselineskip]{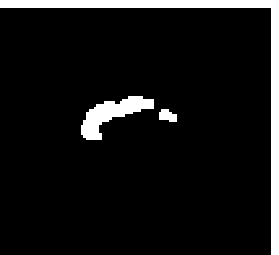}}\hfill%
	\caption{Comparison of segmentation results for different methods, column one is the lesions, second is the ground truth masks and column third, fourth and fifth are the segmentation output for GMM, KM, and MCWT. }
	\label{fig:gull}
\end{figure} 

\section{Discussion and Conclusion}
Lesion segmentation is a crucial step for the characterization of tumors in breast MR images. In this work, we presented a comparison of 3 unsupervised segmentation methods for the task of MRI breast lesions to evaluate their performance. The algorithms have been applied to 106 lesions and MCWT outperformed the other methods marginally. The results presented in Fig.1 shows that MCWT could connect those disjoint areas in the lesion better than other two methods. The markers in watershed transformation typically include the neighborhood pixels which has the lower intensity to a particular region. However, this method is sensitive to noise and the soft edges computed by evaluating the gradient image. KM segmentation approach could not segment some of the pixels within the lesions in comparison to the GMM, as it can be seen in the first two sample cases in Fig.1. 
In general, the key advantage of these unsupervised segmentation methods is that they do not require a manually-segmented reference image. A manually-created ground truth image is intrinsically subjective and creating such a reference image is a time-consuming process, particularly in the case of breast MR lesion segmentation in 3D. Future work will look to extend our proposed 2D watershed algorithm to 3D and combine it with a lesion detection technique, to establish a complete CAD system, with minimum manual intervention.

\bibliographystyle{2554} 

\bibliography{0000}
\end{document}